**RESEARCH PAPER**

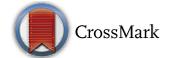

# Design principles for a hybrid intelligence decision support system for business model validation

Dominik Dellermann[1,2] · Nikolaus Lipusch[1,2] · Philipp Ebel[1,2] · Jan Marco Leimeister[1,2]

**Abstract**
One of the most critical tasks for startups is to validate their business model. Therefore, entrepreneurs try to collect information such as feedback from other actors to assess the validity of their assumptions and make decisions. However, previous work on decisional guidance for business model validation provides no solution for the highly uncertain and complex context of early-stage startups. The purpose of this paper is, thus, to develop design principles for a Hybrid Intelligence decision support system (HI-DSS) that combines the complementary capabilities of human and machine intelligence. We follow a design science research approach to design a prototype artifact and a set of design principles. Our study provides prescriptive knowledge for HI-DSS and contributes to previous work on decision support for business models, the applications of complementary strengths of humans and machines for making decisions, and support systems for extremely uncertain decision-making problems.

**JEL classification** D81

✉ Dominik Dellermann
dellermann@uni-kassel.de

Nikolaus Lipusch
lipusch@uni-kassel.de

Philipp Ebel
ph.ebel@uni-kassel.de

Jan Marco Leimeister
leimeister@uni-kassel.de

[1] Research Center for IS Design (ITeG), Information Systems, University of Kassel, Pfannkuchstrasse 1, 34121 Kassel, Germany

[2] Institute of Information Management, University of St. Gallen, Mueller-Friedberg-Strasse 8, 9000 St. Gallen, Switzerland

## Introduction

The rapid digital transformation of businesses and society generates great possibilities for developing novel business models that are highly successful in creating and capturing value. Many Internet startups such as Hybris, Snapchat, and Facebook are achieving major successes and quickly disrupting whole industries. Yet, most early-stage ventures fail. Nearly 90% of technology startups do not survive the first five years (Patel 2015). One reason for this is that entrepreneurs face tremendous uncertainties when creating their business models. Consequently, entrepreneurs must constantly re-evaluate and continuously adapt their business models to succeed (Ojala 2016). This task is characterized by high levels of uncertainty concerning market and technological developments. In addition, entrepreneurs cannot be sure whether their competencies and internal resources are suitable to successfully run the new venture (Andries and Debackere 2007; Timmers 1998). Therefore, entrepreneurs try to collect information that might support them in their decision making. Such information includes the following: analytical data such as market or financial data; feedback from customers and other stakeholders; and guidance from associate mentors, business angels, and incubators. This information is used to assess the validity of their assumptions and make decisions that are necessary to succeed (e.g., Shepherd 2015; Ojala 2016).

However, ways to get decision support in the process of business model validation are limited (Dellermann et al. 2017a). One is the use IT-supported tools to provide guidance for incumbent firms, as shown by some existing research (e.g., Gordijn and Akkermans 2007; Haaker et al. 2017; Daas et al. 2013; Euchner and Ganguly 2014). These tools frequently rely on formal analysis of financial data and forecasts that might work well for established companies. However, the applicability of these approaches in the startup context remains challenging because business ideas are vague, prototypes do not yet exist, and thus the proof of concept is still pending. Moreover, early-stage startups might not have a market yet but offer great potential of growth in the future (Alvarez and Barney 2007). Another way to deal with uncertainty during the validation of a business model is the validation of the entrepreneur's assumptions by testing them in the market or with other stakeholders (Blank 2013). Such validation allows the entrepreneur to gather feedback, test the viability of the current perception of a business model, and adapt it as necessary. This approach includes both formal analysis and the intuition of human experts, which has proven to be a valuable combination in such uncertain settings (Huang and Pearce 2015). For this purpose, traditional offline mentoring is the state of the art in both theory and practice. However, such offline mentoring provides only limited possibilities for scalable and iterative decision support during the design of a business model (e.g., Hochberg 2016).

Therefore, the purpose of this study is to develop a decision support system (DSS) that allows the iterative validation of a business model through combining both social interaction with relevant stakeholders (e.g., partners, investors, mentors, and customers) and formal analysis for the extremely uncertain context of business model development in early-stage startups. Such a combination proved to be most valuable for decisions in this setting (e.g., Huang and Pearce 2015). In particular, we propose a Hybrid Intelligence DSS (HI-DSS) that combines the strength of both machine intelligence to handle large amount of information as well as collective intelligence which uses the intuition and creative potential of individuals while reducing systematic errors through statistical averaging. We follow a design science approach (Hevner 2007; Peffers et al. 2007), making use of both knowledge from previous research that proved to be valuable in various contexts of uncertain decision making, as well as practical insights, to develop principles for an IT artifact.

Our contribution is threefold. First, our research provides prescriptive knowledge that may serve as a blueprint to develop similar DSSs for business model validation in the context of early-stage startups (Gregor and Jones 2007). In fact, our research provides prescriptive knowledge regarding design principles (i.e., form and function) as well as implementation principles (i.e., our proposed implementation). Second, we contribute to research on decision support for business model validation by augmenting formal analysis of data through iterative social interaction with stakeholders. Third, we propose a novel approach to support human decision-making by combining machine and collective intelligence and thus contribute to recent research on ensemble methods (e.g., Nagar and Malone 2011; Brynjolfsson et al. 2016).

# Related work

## Business models and business model validation

To formulate the problem for our design research approach, we reviewed current literature on business model validation. The concept of the *business model* has gathered substantial attention from both academics and practitioners in recent years (Timmers 1998; Veit et al. 2014). In general, it describes the logic of a firm to create and capture value (Al-Debei and Avison 2010; Zott et al. 2011). The business model concept provides a comprehensive approach toward describing how value is created for all engaged stakeholders and the allocation of activities among them (Bharadwaj et al. 2013; Demil et al. 2015). In the context of early-stage startups, business models become particularly relevant as entrepreneurs define their ideas more precisely in terms of how market needs might be served. A business model reflects the assumptions of an entrepreneur and can therefore be considered as a set of "*hypotheses about what customers want, and how an enterprise can best meet those needs and get paid for doing so*" (Teece 2010: 191). Thereby, entrepreneurs make several decisions regarding the design of a business model such as how a revenue model, value proposition, and customer channels should be constructed. Thus, a business model can be used as a framework for constructing startups and to conduct predictive what-if scenario analysis to determine the feasibility of an entrepreneur's current pathway (Morris et al. 2005).

However, such scenario analysis concerning an entrepreneur's assumptions about what might be viable and feasible are mostly myopic in terms of the outcome because entrepreneurs are acting under high levels of uncertainty (Alvarez and Barney 2007). Consequently, entrepreneurs have to start a sensemaking process to gather information for validating and refining their initial beliefs and guiding future decision-making. During this process, the entrepreneurs refine the business model through iterative experimentations and learning from both successful and failed actions. These design decisions determine how a business model is configured along several dimensions (Alvarez et al. 2013; Blank 2013). When the entrepreneurs' assumptions contradict with the reaction of the market, this might lead to a rejection of erroneous hypotheses and require a reassessment of the business model to test the market perceptions again. Thus, the business model evolves toward the needs of the market and changes the assumptions of entrepreneurs (Ojala 2016). The success of startups, thus, depends heavily on the entrepreneurs' ability to develop and continuously adapt their business models to the reactions of the environment by making adequate decisions (Spiegel et al. 2016).

## Decision support for business model validation

Decision support can assist in making business model design decisions (i.e., how a business model should be constructed) in several ways. One, previous research on decision support and validation in the context of business model analysis mainly focuses on analytical methods such as modelling and simulation (e.g., Gordijn et al. 2001; Haaker et al. 2017; Daas et al. 2013; Euchner and Ganguly 2014). Business model simulations provide a time-efficient and cost-efficient way to help decision makers understand the consequences of business model adaptions without requiring extensive organizational changes (Osterwalder et al. 2005). In this vein, previous research applies quantitative scenarios analysis to predict the viability of design decisions in the context of business model innovation for platforms (Zoric 2011) and mobile TV (Pagani 2009), as well as scenario-planning methods for IP-enabled TV business models (Bouwman et al. 2008). Another way of providing guidance in the design of a business model focuses on stochastic analysis of financial models to identify the most important drivers of financial performance in incumbent firms, such as Goodyear (Euchner and Ganguly 2014). A third popular approach evaluates business model design choices against a potential scenario of changes in stress-testing cases (Haaker et al. 2017). Although most of this research considers the importance of the consistency of causal business models structures and the complex interrelations of components, existing methods do not consider how the effects of changes in a business model unfold dynamically over time and the iterative process of developing business models especially for new ventures (Cavalcante et al. 2011; Demil and Lecocq 2010). Most of these approaches are rather static and thus only few are capable of capturing the dynamics that underlie the complex interactions of business model design decisions in practice (e.g. Moellers et al. 2017). Such analytical methods to support decisions in the context of business model validation lack the capability to identify complex pattern of components that lead to success.

While these methods are valuable for incumbent firms, they are not very applicable for early-stage startups. Predicting the success of early-stage ventures´ business models is extremely complex and uncertain. This is due to the fact that neither possible outcomes nor the probability of achieving these outcomes are known, i.e., situations of unknowable risk (Alvarez and Barney 2007). Little data is available and quantifying the probability of certain events remains impossible. In such contexts, formal analysis is a necessary but not sufficient condition to assess if a certain business model design might be viable in the future (Huang and Pearce 2015). In such situations, entrepreneurs pursue two strategies. First, they seek and gather available information that they can process to guide analytical decision-making (e.g., Shepherd et al. 2015). Second, entrepreneurs rely on their experience and gut feeling to make intuition-based decisions. Such intuition has proven to be a valuable strategy for decision-making under uncertainty (Huang and Pearce 2015).

While relying on gut feeling and intuitive decision-making is the purview of successful entrepreneurs during the validation of their business models, we argue that the assessment, processing, and interpretation of additional information that reduces uncertainty and guides decision-making needs support due to the limitations of bounded rationality (March 1978; Simon 1955). This is because entrepreneurs are limited in their capability to access and process information extensively, and therefore not able to optimize their decisions. Moreover, the interpretation of accessed and processed information is constrained by biases and heuristics, frequently leading to bad decisions (Bazerman and Moore 2012; Kahneman 2011; Thaler and Sunstein 2008).

Because entrepreneurial decision makers are constrained by bound rationality, startup mentoring has emerged as strategy to support entrepreneurs in making the required decisions. Mentors (i.e., experienced consultants, experts, or successful entrepreneurs) attempt to help the early-stage startup team to gain problem-solution fit by conducting one-to-one support initiatives (such as workshops) and offer entrepreneurs methods to develop their idea into a novel venture (Cohen and Hochberg 2014). Such social interaction with relevant stakeholders is frequently offered by service providers such as incubators and more recently accelerators (e.g., Cohen and Hochberg 2014; Pauwels et al. 2016). However, neither academia nor managerial practice are offering IT-based solutions to iteratively provide such guidance. This is unfortunate, since IT-based solutions have the potential to provide scalable and cost-efficient solutions by leveraging the wisdom of multiple and diverse mentors, iterate the validation and adaption process, and allow the transference of many entrepreneurs' experiences to a single entrepreneur, thereby increasing the learning rate of the individual entrepreneur.

## Methodology

### Design science research project methodology

Novel solutions are needed to address the limitations of individual decision makers resulting from their bounded rationality and the lack of scalable solutions for providing guidance in business incubators and accelerators. To provide IT-supported forms of guidance to entrepreneurs, we conduct a design science research (DSR) project (Peffers et al. 2007; Gregor and Hevner 2013) to design a new and innovative artifact that helps to solve a real-world problem by providing a high-quality and scalable tool for decision guidance. To conduct our research, we followed the iterative DSR methodology process of Peffers et al. (2007) consisting of six phases: (1) problem identification and motivation, (2) objectives of a solution, (3) design and development, (4) demonstration, (5) evaluation, and (6) communication. We used a multi-step formative ex-ante and summative ex-post evaluation in a naturalistic setting with domain experts and potential users to ensure the validity of our results (Sonnenberg and Vom Brocke 2012; Venable et al. 2016).

Our research starts with phase 1: i.e., the formulation of the problem that is perceived in practice. To ensure both relevance and rigor, we use inputs from the practical problem domain (relevance) and the existing body of knowledge (rigor) for our research project (Hevner 2007). Abstract theoretical knowledge has a dual role. First, it addresses the suggestions for a potential solution. Second, the abstract learning from our design serves as prescriptive knowledge to develop a similar artifact in the future (Gregor and Jones 2007). Therefore, we conducted a literature review on decision support in the context of business model validation. To refine and validate the relevance of this problem, we conducted an explorative study within the problem domain using qualitative interviews (e.g., Dul and Hak 2007; Yin 2013). We collected data concerning the business model validation process within business accelerators and incubators. We conducted a series of expert interviews with executives at business incubators and accelerators ($n = 27$), entrepreneurs ($n = 32$), and mentors ($n = 16$). We gained access to interviewees in the context of a project funded by the German Ministry of Research and Education. Our project partners then helped us with a snow sampling approach to gain access to additional participants in their network. The statements of the interviewees were coded and analyzed by two of the researchers to identify common themes. Our coding procedure was structured and guided by the limitations derived from literature (Strauss and Corbin 1990). This approach allowed us to justify the research gap in practical relevance before designing an artifact (Sonnenberg and Vom Brocke 2012). In a second step, we analyzed previous research on decision support to identify a body of knowledge that provides suggestions for a potential solution resulting in a scientifically grounded version of design principles.

To evaluate our design, we used a combination of exploratory and confirmatory focus groups (Hevner and Chatterjee 2010). Originated in the field of psychology, the focus group gained increasing popularity as a knowledge elicitation technique in the field of software engineering (Massey and Wallace 1991; Nielsen 1997). We used exploratory focus groups to gather feedback for design changes and refinement of the artifact. This was used as formative evaluation procedure to iteratively improve the design. Moreover, a confirmatory focus group was applied to demonstrate the utility of our artifact design in the application domain (Tremblay et al. 2010).

The initial version of the tentative design principles was demonstrated, validated, and refined using eight focus-group workshops (6–8 participants; average duration 60 min) with mentors, executives, and software developer. The design principles were visualized, explained, and discussed to formatively evaluate the completeness, internal consistency, and applicability of our ex-ante design (Sonnenberg and Vom Brocke 2012; Venable et al. 2016). In the next steps, we instantiated the tentative design principles of form and function into an IT artifact. We then applied a summative ex-post evaluation of the design through a qualitative evaluation in a naturalistic setting with potential users. Therefore, we conducted eight focus group workshops with mentors, executives at incubators and corporate accelerators, and entrepreneurs (2–4 participants; average duration: 60 min).[1] The instantiated artifact was explained to the participants and demonstrated by a click-through. The participants then had the possibility to use the artifact and were then asked to assess its effectiveness, efficiency, and fidelity with the real-world phenomenon, which leads to the final version of principles of form and function (Venable et al. 2016). Figure 1 summarizes the six different phases and their activities as they have been conducted in the course of this DSR project. And the following sections elaborate on each of the six phases.

## Artifact description

### Problem verification (phase 1): The validation of early-stage business models in existing mentoring programs

To enable a two-sided perspective on the problem and to ensure the practical relevance of the identified gap (Sonnenberg and Vom Brocke 2012), we conducted a total of 75 exploratory interviews with entrepreneurs and members of incubators. The interviews were guided by the central question of how incubators as service providers typically support the design decisions of entrepreneurs' business models and the perceived limitations of these approaches. Therefore, the interviews were used for understanding the problem domain and gathering deep insights in the real-world phenomenon. By analyzing these interviews, we gained a deeper understanding of the process of business model validation for startups. In sum, it turned out that entrepreneurs face three types of problems when trying to access the quality of their business models. One type involves the bounded rationality of humans that prevents them from extensively searching for the required information and leads to biased decisions. A second type concerns the limitations of current forms of decisional guidance provided to entrepreneurs (e.g., mentoring in business incubators) that prevent the mentors from providing optimal. These limitations include a limited domain of knowledge especially concerning novel technologies, a lack of cognitive flexibility, and the subjectivity of the evaluations. Moreover, resource constraints of institutional mentoring organizations in general make it nearly impossible to find the perfect decisional guidance for each business model case and constrain iterative development. The third type of problems deals with the limitations of existing IT-based tools as discussed in the previous section. Table 1 summarizes the aggregated finding of our literature analysis as discussed in the Related Work section, combined with the findings of the interviews. Based on this, we draw conclusions for consequential objectives.

---
[1] For further details of the problem identification, demonstration, and evaluation see Appendix.

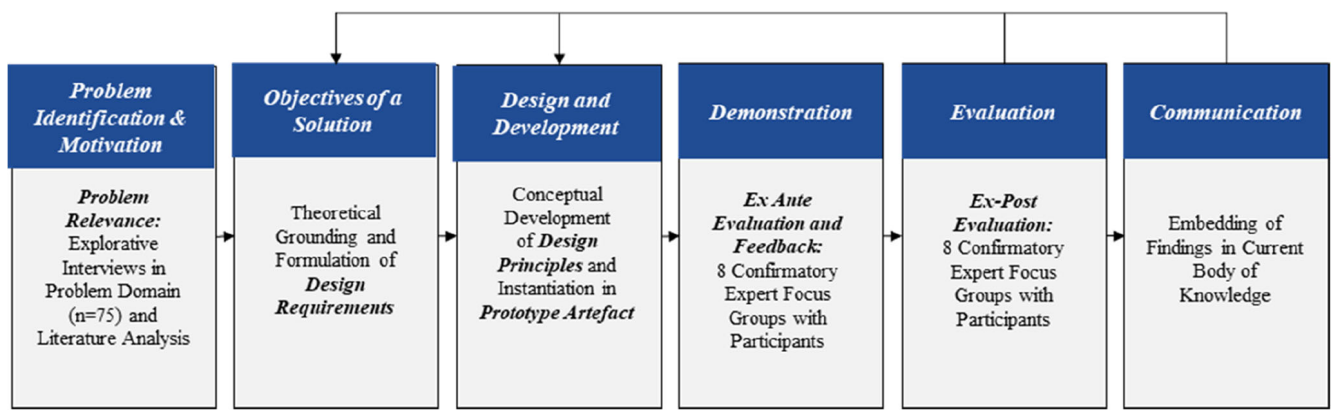

**Fig. 1** Six-phase research process and related procedures (adapted from Peffers et al. (2007))

In sum, the interviews delivered a detailed overview of the problems that entrepreneurs face when trying to assess the quality of their business model. However, we were not able to identify knowledge on how to support this decision process with the help of an IT tool. We therefore investigated existing design theories (i.e., kernel theories) that have been used to solve similar problems (Gregor and Hevner 2013).

## Deriving the objectives of a solution (phase 2)

As outlined, when making decisions regarding their business model, entrepreneurs must improve decision quality to make successful business model design decisions (Sosna et al. 2010). This is obvious as making the most appropriate decisions at a certain point of time is maybe the most challenging task for entrepreneurs (Alvarez et al. 2013). Such design decisions regarding certain business model components can support or prevent the entrepreneur from achieving milestones such as gaining a viable market position or receiving funding (Morris et al. 2005).

In the context of business incubators and other support activities for entrepreneurs, providing guidance through mentoring proved to be the most suitable approach for helping entrepreneurs to design business models. One other domain where decisional guidance has proved to be a suitable approach is research on DSSs in various contexts of IS research (Silver 1991; Morana et al. 2017; Parikh et al. 2001; Limayem and DeSanctis 2000). Such guidance – which supports and offers advice to a person regarding what to do – was examined especially in the context of DSSs (Silver 1991). In this vein, decisional guidance describes design features that enhance the decision-making capabilities of a user (Morana et al. 2017). The adaption of this finding to the context of this research project (i.e., entrepreneurial decision making) is promising.

To support decision makers in executing their tasks, they must be provided with design features of decisional guidance differentiated along ten dimensions (Morana et al. 2017; Silver 1991). First, the target of guidance supports the user in choosing which activity to perform or what choices to make. Second, the invocation style defines how the decisional guidance is accessed by the user, such as automatically, user-invoked, or intelligently (Silver 1991). Third, guidance can be provided at different timings such as during, before, or after the actual activity (Morana et al. 2017). The timing for the context of business model design choices should not be time-specific but rather should be accessible during various time points during the process (Silver 1991). Fourth, decisional guidance can be provided for novices and experts (Gregor and Benbasat 1999). This dimension of design features is also not relevant for our context as entrepreneurs vary in their expertise from novices to experts that are serial entrepreneurs. Fifth, the trust building of decisional guidance is not explicitly covered for business model design decisions (Silver 1991). As this is not an issue in traditional offline mentoring either, we propose that trust issues – though highly relevant for future research – be outside the scope of this paper. Finally, the content dimension of decisional guidance is defined as the purpose of the guidance provision (Morana et al. 2017). As this is highly interrelated with the intention of decisional guidance, we did not explicitly highlight this design feature as a requirement for our solution.

Table 1 Summary of findings of literature analysis and interviews

| Category | Problem | Example | Exemplary References | Consequence |
|---|---|---|---|---|
| Limitations of Entrepreneurs | *Limited Expertise* (e.g. in Technology, Management etc.) | • Lack of Experience | Larrick and Feiler 2015; Shepherd et al. 2015; Alvarez et al. 2013; Blank 2013 | Need for Decisional Guidance |
| | *Bound Rationality* | • Overconfidence<br>• Overoptimism<br>• Illusion of control<br>• Representativeness<br>• Availability | March 1978; Simon 1955; Bazerman and Moore 2012; Kahneman 2011; Tversky and Kahneman 1974; Zhang and Cueto 2017 | |
| Limitations of Current Forms of Guidance | *Limited Expertise* (e.g. in Technology, Management etc.) | • Lack of Experience | Larrick and Feiler 2015; Shepherd et al. 2015; Alvarez and Barney 2007 | Need of IT-based Solutions and Aggregation of Individual Judgements |
| | *Bound Rationality* | • Use of Misleading Heuristics<br>• Similarity Bias<br>• Overconfidence | March 1978; Simon 1955; Bazerman and Moore 2012; Kahneman 2011; Tversky and Kahneman 1974; Zhang and Cueto 2017 | |
| | *Limited Cognitive Flexibility* | • Functional Fixedness | Larrick and Feiler 2015; Zhang and Cueto 2017 | |
| | *Resource Constraints* | • Lack of Access to Suitable Experts<br>• Time Constraints | Cohen and Hochberg 2014; Pauwels et al. 2016 | |
| Limitations of IT-based Tools | *Focus on (Hard) Financial Data and Simulation* | • Financial Forecasts for Start-ups Not Available | Gordijn et al. 2001; Haaker et al. 2017; Daas et al. 2013; Euchner and Ganguly 2014; Gordijn and Akkermans 2007 | Need for Hybrid Intelligence DSS to Capture Extreme Uncertainty and Complex Interactions |
| | *Limited Integration of (Soft) Qualitative Criteria* | • No Use of Soft Factors Such as Team, Innovativeness etc. | Euchner and Ganguly 2014; Dellermann et al. 2017a, b | |
| | *Lack of Capability to Identify Complex Interactions between Pattern* | • No Capturing of Interactions between Business Model Concepts | Moellers et al. 2017 | |

### Deriving objectives from the body of knowledge on decisional guidance

For the scope of this research, we focused on four of these guidance dimensions that are particularly relevant for business model decision support: the form (or directivity), the mode, the intention, and the format of decisional guidance.[2]

Decisional guidance can be divided into two *forms of guidance*: informative guidance, which provides the user with additional information; and suggestive guidance which offers guidance for a suitable course of action. Informative guidance is used as expert advice to provide declarative or definitional knowledge, thereby helping the user to increase the understanding of a decision model (Limayem and DeSanctis 2000). Suggestive guidance makes specific recommendations on how the user should act (Arnold et al. 2004). This form of guidance can improve decision quality and reduce resource requirement for making decisions (Montazemi et al. 1996).

The *mode of guidance* describes how the guidance works. This covers the design feature of how the guidance is generated for the user. It can be predefined by the designer and thus be statically implemented into a system a priori, dynamically learned from the user and generate decisional guidance on demand, and participative depending on the user's input (Silver 1991). Dynamic guidance is particularly useful to improve decision quality, user learning, and decision performance, while participative guidance also increases the performance of users in solving complex tasks (Parikh et al. 2001; Morana et al. 2017).

---

[2] For a comprehensive review of design features for decisional guidance, see Morana et al. (2017).

The *intention of guidance* describes why guidance is provided to users. This might for instance include clarification, knowledge, learning, or recommendation (Arnold et al. 2004; Gönül et al. 2006). Decisional guidance can be provided with the intention to provide specific recommendations on how to act or expert advice to help users solve problems and make decisions.

Finally, the *format of guidance* pertains to the manner of communicating the guidance to the user. Decisional guidance can be formatted for instance as text or multimedia (images, animation, audio) to make it more accessible for the user. The format of guidance be should selected depending of the characteristics of the underlying task (Gregor and Benbasat 1999; Morana et al. 2017).

### Formulation of design requirements

In general, a system that supports the entrepreneur in making design decisions during business model validation needs to support the entrepreneur in executing her task. This requires a certain combination of guidances that are specific for the context (Silver 1991). For the purpose of our research, we structured our design requirements (DR) along the four dimensions that were identified as suitable for the class of decision problem.

First, the form of guidance provided needs to include information about the probability of success of the current version of the business model. This means providing a forecast on the probability of having success in the future such as receiving funding, survival, growth etc.: i.e., *informative guidance* (Morana et al. 2017; Silver 1991). Therefore, we formulate the following design requirement:

> **DR1:** *Business model validation should be supported by a DSS that provides informative guidance to signal the value of the business model.*

Second, the HI-DSS should guide the entrepreneur's adaption of the business model by providing concrete advice on how the elements of the business model should be adapted: i.e., *suggestive guidance* (Montazemi et al. 1996). Therefore, we formulate the following design requirement:

> **DR2:** *Business model validation should be supported by a DSS that provides suggestive guidance to advice on concrete future actions.*

Third, the DSS should learn from the user and generate guidance on user demand as the task of business model validation is highly uncertain and dynamic and does not allow the offering of predefined guidance (**dynamic guidance**). Therefore, we formulate the following design requirement:

> **DR3:** *Business model validation should be supported by a DSS that allows dynamic changes and learns from users' input.*

Fourth, users should be able to determine the guidance provided. In the context of business model validation, this means providing direct guidance through mentors: i.e., **participative guidance**. Both modes of decisional guidance (participative and dynamic) are particularly effective in improving decision quality, user learning, and decision performance in highly complex tasks such as business model validation (e.g., Parikh et al. 2001). Therefore, we formulate the following design requirement:

> **DR4:** *Business model validation should be supported by a DSS that allows participation of users (*i.e. mentors) in providing the guidance.*

Fifth, the DSS should provide additional knowledge to the entrepreneur to give her guidance on how to improve the business model: i.e., **knowledge** (e.g., Schneckenberg et al. 2017). Therefore, we formulate the following design requirement:

> **DR5:** *Business model validation should be supported by a DSS that provides the user with predictive and prescriptive knowledge on the business model.*

Sixth, it is central that the user learns from the actions in the long term: i.e., **learning** (e.g., Alvarez and Barney 2007). Therefore, we formulate the following design requirement:

> **DR6:** *Business model validation should be supported by a DSS that allows the user to learn from iterations.*

In previous studies, combining the above six dimensions of decisional guidance proved to be most suitable when trying to overcome limitations in individual decision-making (Montazemi et al. 1996; Parikh et al. 2001; Mahoney et al. 2003).

Finally, the user needs to properly visualize the decisional guidance in order to be able to draw inferences from it; i.e., **visualization**. Therefore, combining different formats of presenting the results are needed, such as text-based and multimedia (Gregor and Benbasat 1999). The formats should match the characteristics of the task (i.e., business model design decisions) (Vessey 1991). Therefore, we formulate the following design requirement:

> **DR7:** *Business model validation should be supported by a DSS that provides the user with visualization of the guidance.*

## Implementing decisional guidance in highly uncertain contexts

To support decision making, it is essential to provide high-quality guidance to the user (Gregor and Benbasat 1999; Silver 1991; Montazemi et al. 1996). Two recently popular approaches for providing high-quality guidance in decision support in uncertain settings are statistical methods (e.g., Creamer et al. 2016) and collective intelligence (e.g., Surowiecki 2005; Malone et al. 2009).

Computational methods have become particularly valuable due to progress in machine learning and machine intelligence to identify, extract, and process various forms of data from different sources to make predictions in the context of financing decisions (Yuan et al. 2016), financial return (Creamer et al. 2016), and bankruptcy of firms (Olson et al. 2012). Statistical models are unbiased, free of social or affective contingence, able to consistently integrate empirical evidence and weigh them optimally, and not constrained by cognitive resource limitations (Blattberg and Hoch 1990). Machine learning is a paradigm that enables a computer program (i.e., an algorithm) to learn from experience (i.e., data) and thus improves the program's performance (e.g., accuracy) in conducting a certain class of tasks (e.g., classification or regression). Consequently, machine intelligence is capable of making statistical inferences based on patterns identified in previously seen cases and learning as the data input grows (Jordan and Mitchell 2015). In addition, such procedures allow the identification of complex patterns in business model configurations and the interrelation between single components and thus extend methods such as business model simulations and financial scenarios.

Although machine learning techniques might be generally suitable for predicting the consequences of certain business model design choices based on prior data distributions of easily quantifiable features (e.g., firm age, team size), they are often biased and fail in highly uncertain settings, when for instance data shifts over time and the data that was previously used to train the model is no longer representative or patterns emerge that were never seen before by the algorithm (Attenberg et al. 2011; Dellermann et al. 2017b). Furthermore, they are not able to predict the "soft" and subjective evaluations of new ventures such as innovativeness of a value proposition, the vision or the fit of the team, or the overall consistency of a business model, which makes it impossible for machines to annotate such types of data (Cheng and Bernstein 2015; Petersen 2004). Due to these limitations, machine intelligence systems require the augmentation of human intuition to successfully guide the design of business models (Attenberg et al. 2011; Kamar 2016).

Human decision makers bring several benefits. Humans are still the "gold standard" for assessing "data that cannot easily be annotated and trained for machine learning models such as creativity and innovativeness" (Baer and McKool 2014; Cheng and Bernstein 2015). Humans are particularly good at providing subjective judgements of data that is difficult to measure objectively through statistical techniques (e.g. Einhorn 1974; Cheng and Bernstein 2015). Additionally, human experts have highly organized domain knowledge that enables them to recognize and interpret very rare information. Such data might lead to specific outcomes that are difficult to predict a priori and would rather represent outliers in a statistical model (Blattberg and Hoch 1990). Using humans for augmenting machine intelligence proved to be valuable in many other settings (Cheng and Bernstein 2015; Kamar 2016). Pertaining to the context of this research, using human intuition proved to be a valuable strategy for anticipating startup business model success at the early-stage (Huang and Pearce 2015).

While individual humans still have the cognitive limitations discussed in previous chapters, these can be minimized through the mechanism of collective intelligence (Larrick et al. 2011). This approach aggregates the judgments of a larger group of humans to reduce the noise and bias of individual evaluations (Klein and Garcia 2015; Blohm et al. 2016; Leimeister et al. 2009). For making judgments in uncertain settings, the value of crowds over individual experts can be explained by two basic principles: error reduction and knowledge aggregation (Larrick et al. 2011). While an individual decision maker might be prone to biases and errors (such as individual entrepreneurs or mentors in our context), the principle of statistical aggregation minimizes such errors by combining several judgements (Armstrong 2001). Furthermore, aggregating the judgement of several individuals is informative as it aggregates heterogenous knowledge about a certain problem and allows the capture of a fuller understanding of a decision-making problem (Soukhoroukova et al. 2012; Keuschnigg and Ganser 2016; Ebel et al. 2016). Consequently, we argue that collective intelligence represents a proper way to augment machine learning systems by accessing more diverse domain knowledge, integrating it into an algorithm, and reducing the threat of biased interpretation.

Due to these complementary capabilities, we decided to combine machine and collective intelligence for providing decisional guidance to entrepreneurs. We call such combined systems *Hybrid Intelligence Decision Support Systems* (HI-DSS). For the purpose of this research project, we define HI-DSS as a computerized decisional guidance to enhance the outcomes of an individual's decision-making activities by combining the complementary capabilities of human and machines to collectively achieve superior results and continuously improve by learning from each other. HI-DSS might be especially suitable to solve our research problem due to three reasons. First, machines are better at processing analytical information and providing consistent results, especially when data is dispersed and unstructured (Einhorn 1972). In the context of business model validation, this becomes evident through the unstructured data regarding market demands, technological developments, etc. Second, human decision-makers are particularly useful in interpreting and evaluating soft information as they are superior in making judgments about factors like creativity or imagining the future (which are required for start up business models) or providing comprehensive guidance on which action to take (Colton and Wiggins 2012; McCormack and d'Inverno 2014). Third, in highly uncertain and complex situations such as setting up a business model for a new venture, humans can use their intuition and gut feeling which augments statistical methods (Huang and Pearce 2015; Dellermann et al. 2018). In this regard, collective intelligence is applied to capitalize on the benefits of humans while simultaneously minimize the drawbacks of individual decision-makers such as bias or random errors (Larrick et al. 2011).

## Design and development (phase 3)

### Principles of form and function for the hybrid intelligence decision support system

To develop the HI-DSS, we translated the required types of decisional guidance (organized along the conceptually identified dimensions of decisional guidance; Morana et al. 2017) into design principles (principles of form and function) thereby combining mechanisms of machine learning and collective intelligence. This translation process is displayed in Fig. 2.

To apply HI-DSS, entrepreneurs must transfer their implicit assumptions to both human mentors as well as the machine learning algorithm to create a shared understanding. Business models are mental representations of an entrepreneur's individual beliefs that should be made explicit by transferring them into a digital object (Bailey et al. 2012; Carlile 2002). Approaches to transfer such knowledge into a common syntax are required in order to make the knowledge readable for both humans and algorithms (Nonaka and von Krogh 2009). Therefore, ontologies can be used to leverage knowledge sharing through a system of vocabularies. Such ontologies represent popular solutions in the context of business models and include descriptions of a business model's central dimensions, such as value proposition, value creation, and value capture mechanisms (Osterwalder et al. 2005). This allows the user to dynamically and participially provide input in the DSS (**DR3** and **DR4**). Previous work on human cognition has shown that the representation of knowledge in such an object (i.e., digital representation of the business model) should fit the corresponding task (i.e., judging the business model) to enhance the quality of the human guidance (John and Kundisch 2015). Since judging a business model is a complex task, a visual representation is most suitable as it facilitates cognitive procedures to maximize the decision quality (Speier and Morris 2003). This allows human judges to visualize the business model (**DP7**). To make the business model readable for the machine learning algorithm, the ontology requires also a machine-readable format that can be achieved through standardizing the representation of design choices (e.g., pattern) or natural language processing (John 2016). Therefore, we propose the following design principle (DP).

> **DP1:** *Provide the Hybrid Intelligence DSS with an ontology-based representation to transfer an entrepreneur's assumptions and create a shared understanding among the mentors, the machine, and the entrepreneur.*

Past literature shows that a judge who is qualified for providing decisional guidance on business model validation is also an expert in the respective context (Ozer 2009). Such appropriateness results in a higher ability to provide valuable feedback and enables the prediction of the potential future success of a business model even in highly dynamic contexts. Therefore, to be suitable as a judge and provide more accurate guidance, an individual human mentor should have two types of expertise: demand- and supply-side knowledge (Magnusson et al. 2016; Ebel et al. 2016). Demand-type knowledge is necessary to understand the market side of a business model (e.g. Customer, competitors, sales channels, value proposition), which indicates the desirability of a business model. Supple-type knowledge consists of knowledge on feasibility (e.g., resources, activities) and profitability (e.g., cost structure, revenue model) of a business model configuration (Osterwalder et al. 2005; Magnusson et al. 2016). Consequently, it is central for a HI-DSS to match certain business models with specific domain experts to ensure high human guidance quality (**DR3** and **DR4**). Therefore, we propose the following design principle:

> **DP2:** *Provide the Hybrid Intelligence DSS with expertise matching through a recommender system in order that the entrepreneur obtains access to expertise.*

To provide guidance for entrepreneurs, humans need adequate feedback mechanisms to evaluate the developed assumptions (Blohm et al. 2016). From the perspective of behavioral decision-making, this feedback can be categorized as a judgment task in which a finite set of alternatives (i.e., business model dimensions) is evaluated by applying a defined set of criteria by which each alternative is individually assessed by using rating scales. Using multi-attribute rating scales for judging and thus providing informational guidance are most appropriate in this context (Riedl et al. 2013). The multi-criteria rating scales should cover the desirability, feasibility, and profitability of a business model by assessing dimensions which are strong predictors for the future success, such as the market, the business opportunity, the entrepreneurial team, and the resources (Song et al. 2008) (**DR1**). Additionally, the human mentor should be able to provide qualitative feedback to guide the entrepreneur's future action and point towards possible directions. This allows the human to provide additional suggestive guidance (Silver 1991) (**DR2**). Therefore, we propose the following design principle:

> **DP3:** *Provide the Hybrid Intelligence DSS with qualitative and quantitative feedback mechanisms to enable the humans to provide adequate feedback.*

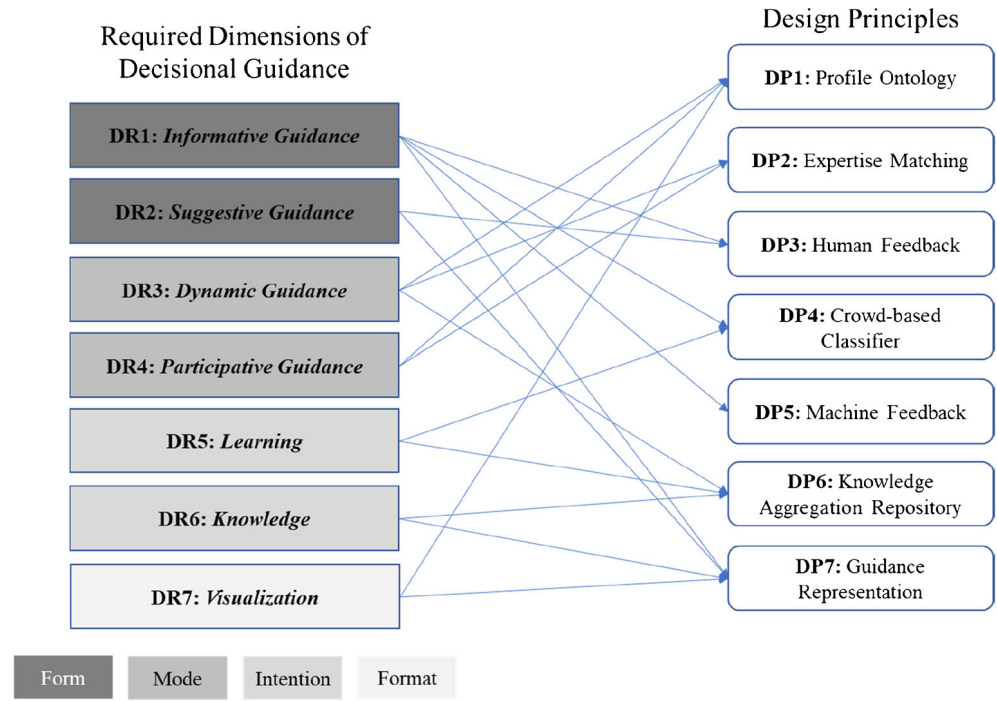

Fig. 2 Mapping design requirements to design principles

In addition to that, the input of the human can also be used to train a machine learning classifier to assess the probability of achieving milestones within the version of the presented business model. This procedure allows consistent processing and weighting of the collective human judgement, which is required to achieve high quality evaluation through collective intelligence (Keuschnigg and Ganser 2016) (**DR1** and **DR4**). Therefore, we propose the following design principle:

> **DP4:** *Provide the Hybrid Intelligence DSS with a crowd-based classifier in order to predict the outcomes of business model design choices based on human assessment.*

Every created business model consists of different design choices (e.g., types of value proposition, revenue models, etc.). This allows supervised machine learning approaches to provide machine feedback concerning the probability of success regarding a certain business model element (e.g., Jordan and Mitchell 2015; John 2016). Supervised algorithms allow a machine to learn from training data (i.e., the user's input) to predict which configuration leads to a favourable outcome (i.e., achieving a milestone) (**DR1**). For this purpose, the user must provide information, so-called "labelling", when a business model achieves a milestone (e.g., funding). This procedure allows training of the machine's ability to evaluate new business model configurations to predict the probability of success of a certain business model version and thus validate (or reject) an entrepreneur's assumptions. Therefore, we propose the following design principle:

> **DP5:** *Provide the Hybrid Intelligence DSS with machine feedback capability in order to predict the outcomes of business model design choices based on statistical assessment.*

Business model validation is an iterative process consisting of validating the existing model, adapting it, and then re-validating it. Therefore, HI-DSS should aggregate the results of each validation round to transient domain knowledge to show how the human and the machine feedback changes an entrepreneur's assumptions and how such changes lead to a certain outcome (e.g., John 2016). The accumulation of such knowledge can trigger cognitive processes that restructure the entrepreneur's understanding of the domain (Sengupta and Abdel-Hamid 1993). Our proposed HI-DSS needs to accumulate in a repository the knowledge created during use, to continuously improve the guidance quality through machine learning, and to learn not only from the iterations of the individual validation process but also from other users of the system (e.g., other entrepreneurs) (Jordan and Mitchell 2015). The knowledge repository can then store general patterns of how changes in a business model are evaluated by humans or the machine learning algorithm, and how they lead to achieving certain milestones (e.g., receiving funding) (**DR3, DR5** and **DR6**). Therefore, we propose the following design principle:

> **DP6:** *Provide the Hybrid Intelligence DSS with a knowledge aggregation repository to allow it to learn from the process.*

Finally, knowledge in the form of additional information and the learning of the entrepreneur must be achieved through a representation of the guidance in a dashboard (Benbasat et al. 1986). Following previous work on business intelligence and decision support visualisations, we argue that the best quality of guidance is achieved when the representation fits the task (e.g. Vessey 1991; Vessey and Galletta 1991). To achieve high decision quality for the user, a cognitive link between the highly complex task (i.e., making business model design decisions) and the guidance should be made by providing visual guidance representation (**DR1, DR2, DR6**, and **DR7**). Moreover, reducing the user's effort for understanding and retrieving the guidance should be achieved by structuring the guidance along the business model dimensions (Baker et al. 2009). Therefore, we propose the following design principle:

> **DP7:** *Provide the Hybrid Intelligence DSS with a visual guidance representation in order that the entrepreneur obtains access to informative and suggestive guidance.*

## Implementation of the hybrid intelligence system[3]

When implementing our HI-DSS, we mapped the identified design principles to concrete design features that represent specific artifact capabilities to address each of the design principles. To implement our design principles into a prototype version of the HI-DSS artifact, we created a cloud-based web-service. The prototype of the artifact consists of a graphic user interface (GUI) that allows the input and visualization of the entrepreneur's business model. For this purpose, we developed a web application in Angular (https://angular.io/). A business model was represented in a standardized and dynamically adaptable format, allowing the entrepreneur to make categorical choices for each element along the value proposition, value delivery, value creation, and value capture dimensions of the business model (Osterwalder and Pigneur 2013) (**DP1**). The expertise matching is achieved through a simple tagging of expertise (i.e., market, technology, or finance). These tags are then matched with an excel that consists of a list of categorized mentors (**DP2**). The feedback mechanisms that allow human judgement are implemented by using the same tool. We implemented 21- criteria Likert rating scales (1 to 10) covering the desirability, implementability, scalability, and profitability of a business model that are commonly applied in practice. Moreover, we provided a textbox for providing concrete qualitative guidance on how to improve the business model. This guidance was structured in terms of the value proposition, value delivery, value creation, and value capture mechanisms of the business model (**DP3**). To gather initial data, we collected publicly available information on startup business models and their respective success to train the machine learning algorithm. The machine learning part of the prototype was developed based on the open source machine learning framework TensorFlow (www.tensorflow.org) in the programming language Python (www.python.org). For the crowd-based classifier we utilized a Classification and Regression Tree (CART) as it provides both good performance and interpretability of results through replication of human decision-making styles (Liaw and Wiener 2002) (**DP4**). We applied the same learning algorithm for analyzing the complex interactions between business model components. Therefore, the success probability of a certain business model is calculated (**DP5**). All the results (i.e., business model components, profile data of mentors, and human judgement) are stored in JSON format in a relational PostgreSQL (www.postgresql.org) database on a Ubuntu SSD server (**DP6**). The final visualization of results (informative and suggestive guidance) is provided through the dashboard implemented in Angular (https://angular.io). This represents aggregated results along the dimensions of desirability, feasibility, and profitability as well as the predicted probability of success along the outcome dimensions of survival and series A funding, which represent commonly accepted proxies for successfully early-stage business models (**DP7**).

The architecture of our prototype artifact is visualized in Fig. 3, while Fig. 4 provides an overview of how the design principles are addressed by design features of the user interface.

---

[3] For access to the prototype please see www.ai.vencortex.com

## Demonstration of the design artifact (phase 4)

The first evaluation of our HI-DSS serves as lightweight and formative ex-ante intervention to ensure that the IT artifact is designed as an effective instrument for solving the underlying research problem (Venable et al. 2016). For this purpose, we decided to make use of exploratory focus groups to refine the artifact design based on feedback from participants. When conducting the focus groups,[4] we followed the process proposed by Tremblay et al. (2010). Within a total of eight focus groups, our design principles were demonstrated, validated, and refined by entrepreneurs and mentors, as well as by developers to validate the technical feasibility of the design principles in a naturalistic setting. During this ex-ante evaluation, we focused on the clarity, completeness, internal consistency, and applicability to solve the practical problem (Sonnenberg and Vom Brocke 2012). The tentative version of the design principles was then adapted before being instantiated into our prototype artifact. The required changes were especially related to the expertise matching and the business model ontology. The participants suggested a switch from well-known business model visualisations (i.e., the business model canvas) to a novel form of startup profiles because mentors require more in-depth information on a certain startup. Especially for a IT-based and time-location asynchronous solution, such an approach is mandatory to ensure high-quality guidance. Moreover, the evaluation revealed the need for expert matching on a fine granular level. Apart from matching industry experts with a startup in a certain domain (e.g., a FinTech startup with a banking industry expert), it is crucial to get feedback from a certain type of expert on each dimension of the business model (e.g., a finance expert for evaluating the value-capture mechanisms). Finally, the participants in the workshop requested the possibility to provide in-depth qualitative feedback to not only point towards suggestions of improvement such as changing the proposed revenue model, but on how to proceed and achieve this goal.

## Evaluation (phase 5)

For the ex-post evaluation of our instantiated design principles into a concrete IT artifact, we applied a qualitative evaluation method to test proof of applicability in the real-world context and to assess the feasibility, effectiveness, efficiency, and reliability against the real-world phenomenon of supporting business model design decisions (Sonnenberg and Vom Brocke 2012; Venable et al. 2016). In particular, we conducted confirmatory focus-group workshops with decision makers and potential users of our HI-DSS in practice. We chose this evaluation approach as a confirmatory method for several reasons. First, the flexibility of the method enabled us to adapt the procedure if necessary. Second, this approach allowed us to directly interact with the potential users of the system, which ensured that the artifact was understood unambiguously. Finally, the focus-group method provided huge amounts of rich data, providing a deeper understanding of the effectiveness and efficiency of the artifact to solve a real-world problem in an actual business environment (Hevner and Chatterjee 2010; Tremblay et al. 2010). For conducting a total of eight focus-group workshops, we recruited 24 participants from business incubators and accelerators as well as independent start-up mentors. Four of the focus groups consisted of participants from business incubators, two from accelerators and three with independent mentors. We presented the HI-DSS via a click-through approach and explained each of the design principles in detail. Then the workshop was guided by the effectiveness of solving the real-world problems and the identified research gap.

As the results of our evaluation show, the HI-DSS overcomes the limitations of previous solutions by combining the analytical processing of interaction between complex business model patterns and the input provided by human intuition. The focus groups reveal that decisional guidance in general helps entrepreneurs to deal with the highly complex and uncertain task of making business model design decisions and overcoming their individual limitations. Moreover, a IT-based solution that aggregates the collective judgement of individual mentors allows for reducing subjectivity while aggregating knowledge that can be stored through machine learning. Finally, the hybrid nature of our proposed design allows it to deal with soft factors and extreme uncertainty by having human intuition in the loop. In addition, using machine learning to identify the complex interaction between different business model elements allows it to deal with the complexity of startup business models.

We then continued the summative ex-post evaluation by assessing each design principle in detail. All the proposed design principles and their instantiations were perceived as useful and effective in solving the problems that are faced during the task execution of decision support for business model validation. The participants argued that the HI-DSS is particularly suitable to improving decision quality and efficiency of the entrepreneurs (**DR6: *knowledge***) and helping them to learn (**DR5: *learning***). The digital nature of the tool was perceived as saving time and resources, and allowed mentors to provide guidance independent of time and location, thus providing high-quality guidance (**DR1: *informative*** and **DR2: *suggestive***). The executives of business incubators and accelerators praised the possibility of accumulating knowledge on the business model design of startups and the implicit sharing of such knowledge through the machine learning approach (**DR3:** *dynamic*). The experts agree that this might increase the survival rate of new ventures at an early stage. The participants liked the possibility to use external mentors from different industries. While the public-funded incubators evaluated the applicability in this context as very high, profit-oriented accelerator mentioned that compensation methods for external mentors beyond intrinsic rewards should be defined. Although altruistic mentoring works well in practice (e.g., business plan competitions and feedback), reward mechanism should be considered to apply our artifact in practice. Furthermore, the experts see great potential in improving decisional guidance through machine learning. They indicate that due to the human component of the HI-DSS, acceptance of the guidance might be higher among entrepreneurs than with only statistical modelling and simulations (**DR4:** *participative* guidance). Finally, the dashboard for visualizing the decisional guidance through graphs and feedback text was perceived as favourable to make the decisional guidance easily accessible for entrepreneurs (**DR7:** *visualization*).

---
[4] For further details of the problem identification, demonstration, and evaluation phases, see Appendix.

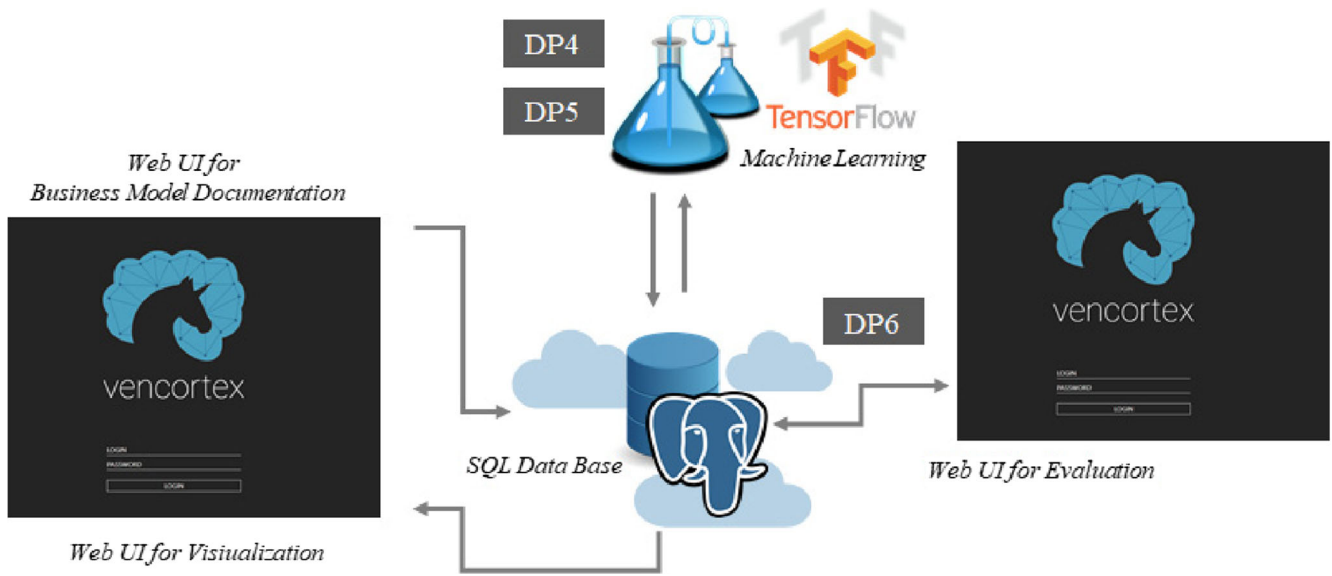

Fig. 3 Architecture of prototype artifact

However, the results of the evaluation also reveal two criticisms that should be resolved before use in a real-life setting. First, the participants highlighted the need for creating trust in AI-based DSSs. While providing highly accurate decisional guidance is crucial, there is a trade-off between accuracy and transparency, which was highlighted by most of the participants. Future research could examine this issue when applying HI-DSS in business contexts. Second, the participants indicated that such IT-based guidance might be perceived as missing the in-depth support of personal mentors. Although the value of the HI-DSS was obvious for all participants, they argued that for communicating with the users of such systems (i.e., entrepreneurs and mentors), the human should still be the focus, while augmented by machine intelligence.

## Discussion

Our results imply that the complementary capabilities of formal analysis and pattern recognition provided through machine learning combined with human intuition provided through collective intelligence is a valuable solution to the extremely uncertain context of iterative business model validation in early-stage startups. HI-DSS enables mentors to provide the required decisional guidance to support entrepreneurs in making their decisions. This is due to several reasons.

First, our proposed design principles capture complex interaction between business model design decisions and the dynamic nature of such choices, and thus overcome limitations of traditional analytical methods such as modelling and simulation (e.g., Haaker et al. 2017; Euchner and Ganguly 2014). Our findings, thereby, also provide a novel and innovative approach in line with previous research on dynamics and complex interactions (e.g., Moellers et al. 2017).

Second, the HI-DSS augments traditional analytical methods to human intuition. By leveraging collective intelligence rather than individual decision makers, our approach prevents the limitations of individual mentors. Consequently, the HI-DSS benefits from the heterogenous knowledge of several experts and aggregates the evaluations of a larger group for reducing the noise and bias of individual judgements. This procedure is particularly valuable in the uncertain and complex context of supporting business model design decisions, providing not only informative guidance in form of business model evaluations but also suggestive guidance

**Fig. 4** GUI of the hybrid intelligence DSS

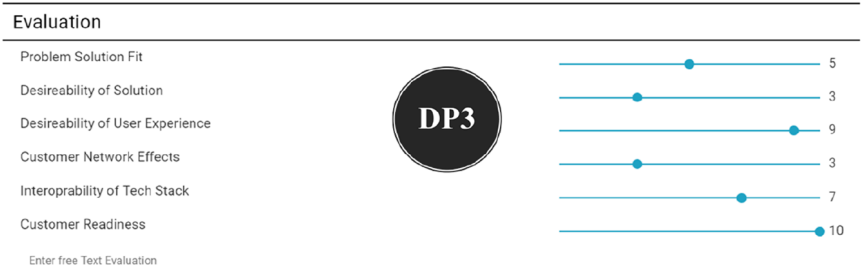

that points entrepreneurs towards direct interventions to improve the business model.

Third, the HI-DSS stores the created knowledge in a knowledge repository. In the long run, this allows both entrepreneurs and mentors to learn from the experience of others. In the course of its use, the system learns what business model design decisions are evaluated positively and negatively by humans and which decisions lead to a specific outcome in a certain context. This may allow the full automation of such decisional guidance in the future.

Fourth, the digital nature of the HI-DSS provides a way to digitize human mentoring. Such IT systems can iteratively validate a business model as well as provide asynchrony and location-independent feedback for resource-efficient mentoring.

Finally, our results point towards a new class of DSS that might be particularly valuable in highly uncertain contexts. With increasing uncertainty, the relative advantages of statistical methods in providing decisional guidance decrease and the value of human intuition increases. As the combinatory nature of formal analysis and intuition during predictions in extremely uncertain contexts is commonly accepted (e.g., Huang and Pearce 2015), such HI-DSS can provide high-quality guidance that might also work in different settings such as innovation or medicine. Our proposed design principles provide a first step in this direction.

Figure 5 depicts the proposed design principles for a HI-DSS in a schematic visualization of the workflow used in the HI-DSS.

## Conclusion

Determining business models for startups is a highly challenging and uncertain task for entrepreneurs and requires various decisions regarding the design of the business model. Due to limitations of individual human decision-makers, this process is frequently tainted by poor decision making, leading to substantive consequences and sometimes even failure of the new venture. As most DSS for business model validation rely on simulations or modelling rather than human intuition, there is an obvious gap in literature on such systems.

Using DSR project methodology, we analysed problems in making decisions about business model design in uncertain environments. We then developed and refined design principles for a HI-DSS that combines the specific benefits of machine and collective human intelligence to steer entrepreneurial decision making by providing decisional guidance. We then instantiated our design principles into a prototype artifact and evaluated them using several focus-group workshops with domain experts.

## Contributions

Our study makes several contributions, both theoretical and practical. First, our research provides prescriptive knowledge that may serve as a blueprint to develop similar DSSs for business model validation (Gregor and Jones 2007). In particular, the findings of this paper reveal prescriptive knowledge about form and function (i.e., design principles) as well principles of implementation (i.e., our proposed instantiation). Due to utilizing justificatory knowledge from the body of knowledge on decisional guidance and the justification of the research gap in both theory and practice, we provide meaningful interventions in the form of design principles to solve a real-world problem and contribute to the discussion of decision support in business model validation.

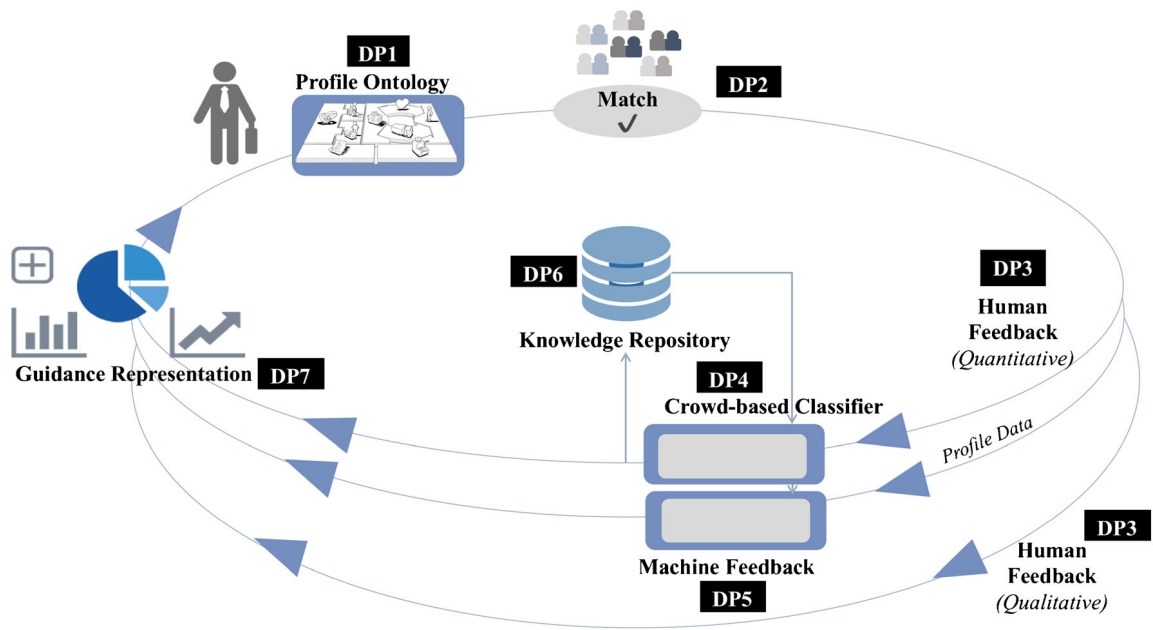

**Fig. 5** Summary of design principles

Second, our results indicate a possible application of collective intelligence in more complex and knowledge-intensive tasks. While previous work (e.g., Blohm et al. 2016; Klein and Garcia 2015) utilized the wisdom of the crowd in rather basic decision support settings such as filtering novel product ideas without considering explicit expertise requirements, our findings indicate the potential of applying collective intelligence in uncertain decision tasks. Addressing the concrete expertise requirements of humans, decisional guidance is based on the heterogenous domain knowledge of experts and reduces misleading biases and heuristics.

Third, we propose a novel approach to support human decision-making by combining machine and collective intelligence into a hybrid intelligence system. Our results show that this form of decisional guidance is particularly relevant in situations of extreme uncertainty where a combination of formal analysis through machine learning techniques and human intuition through collective intelligence is most valuable. Thus, our research contributes to recent work on combined applications in different domains (e.g., Nagar and Malone 2011; Brynjolfsson et al. 2016).

Fourth, we contribute to research on decision support for business model validation by augmenting formal analysis of data to iterative social interaction with stakeholders (e.g., Gordijn et al. 2001; Haaker et al. 2017; Daas et al. 2013; Euchner and Ganguly 2014). This research takes human guidance and judgement into account to help decision makers to design business models. Moreover, the findings start a novel discussion in the field of research on DSS: how can such systems be designed for situations of extreme uncertainty where no objective truth exists.

Finally, our proposed prototype artifact offers an actual solution for helping service providers such as business incubators and accelerators to extend their service offering beyond solely offline mentoring to a digital solution and thus provides a first step towards a practical solution in this context. Based on the results of this paper, further research is focusing on the provision of Hybrid Intelligence services in real-world applications.[5]

## Limitations and further work

Despite its various contributions to theory and practice, our work is not without limitations. First, we focused our research on the context of business incubator and accelerators to provide a DSS that helps them to provide decisional guidance to entrepreneurs. This setting implies that access to a network of mentors is already available and that advise is mainly offered with altruistic motives. Such DSS might require adaption for attracting experts to participate and provide advise via the system. Therefore, further research might explore the motives of such mentors and how DSS might be extended through activation supporting components (e.g., Leimeister et al. 2009). Second, we chose a qualitative evaluation procedure to assess the applicability and effectiveness of a HI-DSS in providing decision support for the business model design process. Although we intended to evaluate in a naturalistic setting with potential users and domain experts, our evaluation procedures were not capable of testing the actual quality of guidance provided by the HI-DSS or its value during long-term use. Further research might therefore develop hybrid prediction algorithms to evaluate the performance (e.g., accuracy) of HI-DSS, particularly compared to other methods. Moreover, a longitudinal study of the use of a HI-DSS in a real-world context might be useful for determining the value of such a system. Finally, our study is limited to the field of business model validation for startups. However, it starts a discussion on a valuable novel form of DSS that combines humans and machines, and as such, encourages exploration of HI-DSS applicability in other settings of uncertainty such as medicine, job applications, and innovation contexts.

---

[5] For further information please see www.vencortex.com